\def\isarxiv{1} 

\ifdefined\isarxiv
\documentclass[11pt]{article}

\usepackage[numbers]{natbib}

\else
\documentclass{article}
\usepackage{neurips_2022}
\fi

\usepackage{amsmath}
\usepackage{amsthm}
\usepackage{amssymb}
\usepackage{algorithm}
\usepackage{subfig}
\usepackage{algpseudocode}
\usepackage{graphicx}
\usepackage{grffile}
\usepackage{wrapfig,epsfig}
\usepackage{url}
\usepackage{xcolor}
\usepackage{epstopdf}

\usepackage{bbm}
\usepackage{dsfont}

\allowdisplaybreaks

\ifdefined\isarxiv

\usepackage{tikz}
\usepackage{hyperref}  
\hypersetup{colorlinks=true,citecolor=blue,linkcolor=blue} 
\usetikzlibrary{arrows}
\usepackage[margin=1in]{geometry}

\else

\usepackage{microtype}
\usepackage{hyperref}
\definecolor{mydarkblue}{rgb}{0,0.08,0.45}
\hypersetup{colorlinks=true, citecolor=mydarkblue,linkcolor=mydarkblue}

\fi

\newtheorem{theorem}{Theorem}[section]
\newtheorem{lemma}[theorem]{Lemma}
\newtheorem{definition}[theorem]{Definition}

\newtheorem{corollary}[theorem]{Corollary}

\newtheorem{assumption}[theorem]{Assumption}

\newtheorem{fact}[theorem]{Fact}
\newtheorem{remark}[theorem]{Remark}

\newcommand{\wh}{\widehat}
\newcommand{\wt}{\widetilde}

\newcommand{\N}{\mathcal{N}}
\newcommand{\R}{\mathbb{R}}

\renewcommand{\hat}{\wh}

\DeclareMathOperator*{\E}{{\mathbb{E}}}

\DeclareMathOperator{\poly}{poly}

\DeclareMathOperator{\nnz}{nnz}

\DeclareMathOperator{\rank}{rank}
\DeclareMathOperator{\diag}{diag}

\DeclareMathOperator{\tr}{tr}

\DeclareMathOperator{\reg}{reg}

\makeatletter
\newcommand*{\RN}[1]{\expandafter\@slowromancap\romannumeral #1@}
\makeatother

\newcommand{\Yichuan}[1]{{\color{blue}[Yichuan: #1]}} 

\usepackage{lineno}

\begin{document}

\ifdefined\isarxiv

\date{}

\title{Zero-th Order Algorithm for Softmax Attention Optimization}

\author{
Yichuan Deng\thanks{\texttt{ycdeng@cs.washington.edu}. The University of Washington.}
\and
Zhihang Li\thanks{\texttt{lizhihangdll@gmail.com}. Huazhong Agricultural University.}
\and 
Sridhar Mahadevan\thanks{\texttt{smahadev@adobe.com}. Adobe Research}
\and
Zhao Song\thanks{\texttt{zsong@adobe.com}. Adobe Research.}
}

\else

\title{Intern Project}  
\maketitle 
\fi

\ifdefined\isarxiv
\begin{titlepage}
  \maketitle
  \begin{abstract}
Large language models (LLMs) have brought about significant transformations in human society. Among the crucial computations in LLMs, the softmax unit holds great importance. Its helps the model generating a probability distribution on potential subsequent words or phrases, considering a series of input words. By utilizing this distribution, the model selects the most probable next word or phrase, based on the assigned probabilities. The softmax unit assumes a vital function in LLM training as it facilitates learning from data through the adjustment of neural network weights and biases. 

With the development of the size of LLMs, computing the gradient becomes expensive. However, Zero-th Order method can approximately compute the gradient with only forward passes. In this paper, we present a Zero-th Order algorithm specifically tailored for Softmax optimization. We demonstrate the convergence of our algorithm, highlighting its effectiveness in efficiently computing gradients for large-scale LLMs. By leveraging the Zeroth-Order method, our work contributes to the advancement of optimization techniques in the context of complex language models. 

  \end{abstract}
  \thispagestyle{empty}
\end{titlepage}

\newpage

\else

\begin{abstract}
Large language models (LLMs) have brought about significant transformations in human society. Among the crucial computations in LLMs, the softmax unit holds great importance. Its helps the model generating a probability distribution on potential subsequent words or phrases, considering a series of input words. By utilizing this distribution, the model selects the most probable next word or phrase, based on the assigned probabilities. The softmax unit assumes a vital function in LLM training as it facilitates learning from data through the adjustment of neural network weights and biases. 

With the development of the size of LLMs, computing the gradient becomes expensive. However, Zero-th Order method can approximately compute the gradient with only forward passes. In this paper, we present a Zero-th Order algorithm specifically tailored for Softmax optimization. We demonstrate the convergence of our algorithm, highlighting its effectiveness in efficiently computing gradients for large-scale LLMs. By leveraging the Zeroth-Order method, our work contributes to the advancement of optimization techniques in the context of complex language models. 
\end{abstract}

\fi

\section{Introduction}

In the last few years, the field of natural language processing has witnessed explosive growth in large language models (LLMs). A series of breakthrough neural network models have rapidly advanced the capabilities of LLMs, including Transformer \cite{vsp+17}, GPT-1 \cite{rns+18}, BERT \cite{dclt18}, GPT-2 \cite{rwc+19}, GPT-3 \cite{bmr+20}, PaLM \cite{cnd+22}, OPT \cite{zrg+22}. Each iteration incorporates architectural innovations and larger datasets to push the boundaries of what is possible with self-supervised learning on text. The conversational chatbot ChatGPT \cite{cha22} created by OpenAI in 2022 brought LLMs into the public spotlight by showcasing their potential for remarkably human-like interaction. Riding this wave, OpenAI recently unveiled an even more powerful LLM called GPT-4 \cite{o23}. While technical details remain scarce, initial evaluations suggest GPT-4 significantly outperforms its predecessor ChatGPT  \cite{bce+23}. 
Fine-tuned LLMs have proven adept at real-world natural language tasks including machine translation \cite{hwl21}, sentiment analysis  \cite{uas20}, language modeling \cite{mms+19}, and even creative writing \cite{cha22,o23}. The rapid progress shows the power of scale and self-supervision in language models. 

\emph{Attention mechanism} is a crucial component of large language models (LLMs) like GPT-3, enabling them to focus on relevant parts of the input text \cite{vsp+17, rns+18, dclt18, rwc+19, bmr+20}. The attention matrix represents correlations between tokens, with entries quantifying the relevance of each token to others. This allows selective focus on pertinent input when generating output, rather than weighing all tokens equally. Attention is inspired by how humans pay differing amounts of attention to various input stimuli. In LLMs, attention is commonly implemented via soft weighting using the softmax function. The attention computation proceeds as follows \cite{zhdk23,as23,bsz23},

\begin{definition}[Static Attention Computation]
    Let $Q,K, V \in \R^{n \times d}$ be three matrices, we define two matrices
    \begin{align*}
        A :=&~ \exp(Q K^\top) \in \R^{n \times n} \\
        D :=&~ \diag( A {\bf 1}_n ) \in \R^{n \times n}
    \end{align*}
    Obviously, $A$ is square and $D$ is a diagonal matrix. Based on these, we define
    \begin{align*}
        \mathsf{Att}(Q,K,V) := D^{-1} A V
    \end{align*}   
    Here ${\bf 1}_n \in \R^n$ is a length-$n$ vector where all the entries are ones.
\end{definition}
In the provided definition, the query tokens, represented by the matrix $Q \in \mathbb{R}^{n \times d}$, are commonly derived from the decoder's preceding hidden state. As for the key tokens and values, we utilize matrices $K \in \mathbb{R}^{n \times d}$ and $V \in \mathbb{R}^{n \times d}$ respectively. The attention matrix $A$ is computed as follows: we take the dot product between each query vector $q_i$ and key vector $k_j$ to obtain the relevance scores, and then apply the softmax function to normalize these scores into attention weights $A_{i,j}$. Specifically,
\begin{align*}
    A_{i,j} = \mathrm{softmax}(q_i^{\top} k_j)
\end{align*}
So each entry $A_{i,j}$ reflects how much attention should be placed on the $j^\text{th}$ key when interpreting the $i^\text{th}$ query token. This enables the model to concentrate on relevant parts of the keys for each query. 

Motivated by the exponential function used in attention, some work has explored hyperbolic regression problems for examples $f(x) = \exp(Ax), \cosh(Ax), \sinh(Ax)$ \cite{lsz23,gms23}, formally defined as follows,
\begin{definition}[Hyperbolic Regression \cite{lsz23}]
    Let $A \in \R^{n \times d}$ and $b \in \R^n$ be a matrix and a vector, we define the objective function of hyperbolic regression problem as
    \begin{align*}
    \min_{x \in \R^d} \| f(x) - b \|_2^2 .
    \end{align*}
    In this case, the function $f(x)$ can take the form of either $\exp(Ax)$, $\cosh(Ax)$, or $\sinh(Ax)$. 
\end{definition}
Very recently, \cite{dls23} considered the normalization factor, and defined the following Softmax regression problem,
\begin{definition}[Softmax Regression, \cite{dls23}]
    Let $A \in \R^{n \times d}$ and $b \in \R^n$ be a matrix and a vector,  we define the objective function of softmax regression problem as
    \begin{align*}
    \min_{x \in \R^d} \| \langle \exp(Ax) , {\bf 1}_n \rangle^{-1} \exp(Ax) - b \|_2^2 .
    \end{align*}
\end{definition}
While in practice of LLMs, the number of parameters to be trained is very large (e.g. ChatGPT has 1.5B parameters \cite{cha22}), training can be explosively slow. A traditional method to avoid this is the \emph{Zero-th Order} methods. A widely-used zero-th order method is the following simultaneous perturbation stochastic approximation (SPSA) \cite{Spa92,spa98} algorithm. 
\begin{definition}[Simultaneous Perturbation Stochastic Approximation (SPSA) \cite{Spa92}]
\label{def:spsa_intro}
    Let $L(x)$ be a loss function. For a point $x_0 \in \R^d$, we define the Simultaneous Perturbation Stochastic Approximation (SPSA) of $L(x)$ on $x_0$ as a vector $\wh{g}(x_0) \in \R^d$ such that
    \begin{align*}
        \wh{g}(x_0)_{i} := \frac{1}{2\epsilon \cdot p_i} ( L(x_0 + \epsilon\cdot p) - L(x_0 - \epsilon\cdot p) ), ~~\forall i \in [d],
    \end{align*}
    where $p \in \R^d \sim \mathcal{N}(0, I_d)$ is the perturbation vector and $\epsilon > 0$ is the perturbation scale. 
\end{definition}
In SPSA, the gradient is approximated using only loss function evaluations, rather than backpropagation. Specifically, random perturbations are added to the parameters, and the loss is evaluated twice - once with positive perturbations, and once with negative. The gradient is estimated as the difference in losses divided by the perturbation size. This allows gradient estimation without explicit differentiation, enabling efficient training of massive models \cite{mgn+23}. While not as accurate as true gradients, SPSA gradients are much cheaper to obtain.

\subsection{Our main result}
In this work, we consider the following loss function: 
\begin{definition}[Our Softmax Loss Function]
\label{def:softmax_loss_informal}
For a vector $x \in \R^d$, we define the softmax loss function 
\begin{align*}
    L_{\exp}(x):= \sum_{j=1}^n L_{\exp,j}(x), ~~~~
    L(x) := \sum_{j=1}^n L_{\exp,\reg,j}(x)
\end{align*}
where
\begin{align*}
    L_{\exp,j}(x) :=& 0.5 \| \langle \exp(A_j x) , {\bf 1}_n \rangle^{-1} \exp(A_j x) - b_j \|_2^2 \\
    L_{\exp,\reg,j}(x) :=& 0.5 \| \langle \exp(A_j x) , {\bf 1}_n \rangle^{-1} \exp(A_j x) - b_j \|_2^2 + 0.5 \| W A_j x\|_2^2
\end{align*}
$A_j \in \R^{n \times d}$, $b_j \in \R^n$. For a certain batch $\mathcal{B} \in [n]$ of data points, we define
\begin{align*}
    L_{\exp}(x;\mathcal{B}) := \sum_{j \in \mathcal{B}} L_{\exp,j}(x).
\end{align*}
\end{definition}
With the experiments of applying SPSA on LLMs \cite{mgn+23}, we look for the underlying theoretical explanation of the performance of SPSA on the large models. We show that, 
\begin{theorem}[Informal version of Theorem~\ref{thm:global_conv_formal}]
\label{thm:global_conv_informal}
     Let $A_j \in \R^{n \times d}$,  Let $b_j \in \R^n$ satisfy that $\| b_j \|_1 \leq 1$ for all $j \in [n]$. Let $R \geq 4$, $\| A_j \| \leq R$, $\| x \|_2 \leq R$, let $M := \exp( O(R^2 + \log n) )$. Let $W = \diag(w)$, where $\min_i w_i^2 \geq {\mu}/{\sigma_{\min}(A_j)}$ for all $j \in [n]$, $|\mathcal{B}| = B$ let $\kappa(A)=\max_{j\in [n]} \kappa(A_j)$.  Let $T=O( M \cdot(1 + d^{1.5} \cdot \kappa^2(A) / k  ) \cdot   \mu^{-2} B^{-1} \log ( ( L(x_0) - L^*) / \epsilon ) )$. Let $x^0$ denote the init point of SGD. Let $L^*=\min_x L(x)$.
    The SGD based on zero-th order method on multiple softmax loss function converges to optimal with an additive error $\epsilon$ in $T$
    iterations. 
\end{theorem}

\subsection{Related work}
\paragraph{Attention Theory.}
Much research has examined the theory behind attention computation in large language models \cite{kkl20, szks21, clp+21, gms23, lsz23, llr23, zhdk23, as23, bsz23, lsz23, dms23, dls23, gsx23, lsx+23, gsy23_hyper, gsy23_dp, zsz+23, wyw+23, ssz23, gms23}. This includes work on approximation algorithms to reduce complexity, like locality sensitive hashing \cite{kkl20, clp+21, zhdk23} and KDEformer \cite{zhdk23}, and analysis of static versus dynamic attention \cite{as23, bsz23}. Hyperbolic regression problems involving exponential and trigonometric activations have also been studied to improve extrapolation \cite{lsz23}. \cite{dls23} studied the softmax regression inspired by attention models. To explain models' attention to salient words and the evolution of attention during training, \cite{szks21} investigated Knowledge to Translate Individual Words (KTIW). They showed KTIW is first learned from word co-occurrence statistics, then later guides attention to relevant input words for output prediction. \cite{llr23} investigated how transformers capture word co-occurrence patterns. Through experiments and mathematical analysis examining network size, depth, and complexity, they showed the embedding and self-attention layers learn to encode topical structure. This was evidenced by higher average inner product and attention weights between words of the same topic. Overall, attention mechanisms have spawned rich theoretical analysis from multiple perspectives.

\paragraph{Zero-th Order Optimization.}
Zero-th Order method is widely used to approximate the gradient. Zero-th order (ZO) or derivative-free optimization methods have a long history in the optimization literature \cite{kw52, spa87, nm65}. These methods estimate gradients using only function evaluations, without explicit derivatives. The two-point gradient approximation dates back to the Kiefer-Wolfowitz algorithm in the 1950s \cite{kw52}. Spall subsequently proposed the simultaneous perturbation stochastic approximation (SPSA) method \cite{spa87}, which uses simultaneous random perturbations to estimate gradients. Another early ZO technique is the Nelder-Mead simplex algorithm \cite{nm65}. More recently, ZO methods have gained popularity in machine learning to handle nonsmooth objectives \cite{djww15}, constraint black-box models \cite{bg22}, and situations where gradients are unavailable or expensive \cite{lrv+20}. Key applications include adversarial attack generation \cite{czs+17}, hyperparameter tuning \cite{sla12}, and reinforcement learning \cite{shc17}. Aforementioned simultaneous perturbation stochastic approximation (SPSA) \cite{Spa92, spa98, mgn+23} is a notable kind of zero-th order optimization method.
Representative ZO algorithms include ZO gradient descent \cite{ns17}, ZO-SGD \cite{gl13, mgn+23, zhl+23}, ZO sign descent \cite{lcch18}, and ZO Frank-Wolfe \cite{bg22}. There are also some zero-th order method which can optimize without no estimation on the gradient \cite{mgr18, gkk+19, h22}. Overall, ZO optimization is crucial when first-order derivatives are infeasible to obtain, while still allowing gradient-based-like algorithms.
Recently, \cite{mgn+23} provide a variant of SPSA algorithm, with less memory use. They also provide experiments on LLMs to show the efficiency of their algorithm. Later \cite{zhl+23} expanded the work \cite{mgn+23}. They conducted distributed fine-tuning with low bandwidth, by the idea of shared randomness. 

\section{Preliminary}
In this section, we state preliminary for the whole paper. 
In Section~\ref{sec:preli:notations}, we define the notations to be used in the paper. 
In Section~\ref{sec:preli:s_rank_and_e_rank} we provide the definition for stable rank and effective rank.
In Section~\ref{sec:preli:algebra} we state a standard tool for exact computation.
In Section~\ref{sec:preli:matrix_norm} we provide some basic tools for matrix norm bounds.
In Section~\ref{sec:preli:psd} we provide some basic tools for matrix inequality.
In Section~\ref{sec:preli:definitions} we define the definitions to be used in our paper.
In Section~\ref{sec:general_defs} we define some basic definition regarding to a function's properties.
In Section~\ref{sec:preli:spsa} we privode the definition of Simultaneous Perturbation Stochastic Approximation(SPSA).
In Section~\ref{sec:preli:previous} we state some previous results from previous work to be used in our paper.

\subsection{Notations}\label{sec:preli:notations}
In this paper, we use $\R$ to denote real numbers, $\R_{\geq 0}$ to denote non-negative real numbers.

Given vector $x \in \R^d$, we $b = \exp(x) \in \R^d$ to generate a vector such that $b_i = \exp(x_i)$ where $i \in [d]$

Given $x \in \R^n$, its $\ell_2$-norm can be denote as $\| x \|_2 := ( \sum_{i=1}^n x_i^2 )^{1/2}$.

Given $A \in \R^{n \times k}$, its spectral norm can be denote as $\| A \|$, i.e.$\| A \| := \sup_{x\in\R^k} \| A x \|_2 / \| x \|_2$.

Given $A$, its largest singular value is denoted as $\sigma_{\max}(A)$, its smallest singular value is denoted as $\sigma_{\min}(A)$.

Given $x \in \R^n$, we use $\| x \|_{\infty}$ to denote $\max_{i \in [n]} |x_i|$.

Given $x \in \R^n, y \in \R^n$, we use $c = x\circ y$ to generate a vector $c \in \R^n$ where $c_i = x_i y_i$ for $i \in [n]$.

Given $x \in \R^n$, we use $A = \diag(x) \in \R^{n \times n}$ to denote a diagonal matrix where $A_{i,i} = x_i$ for $i \in [n]$.

We use $a = {\bf 1}_d \in \R^d$ to denote a vector such that $a_i = 1$ where $i \in [d]$

Given $A,B \in \R^{d \times d}$, we say $A \succeq B$ if $x^\top  Ax \geq x^\top B x$ for $\forall x \in \R^d$.

We define $\cosh(x) = \frac{1}{2}( \exp(x) + \exp(-x))$ and $\sinh(x) = \frac{1}{2} ( \exp(x) - \exp(-x))$.

Given $A \in \R^{n \times d}$, we define the number of non zero entries of $A$ to be $\nnz(A)$, i.e., $\nnz(A) := | \{ (i,j) \in [n] \times [d] ~|~ A_{i,j} \neq 0 \} |$

Given diagonal matrix $D \in \R^{n \times n}$, we say $D$ is a $k$-sparse diagonal matrix where $k := |\{ i \in [n] ~|~ D_{i,i} \neq 0 \}|$.

Given function $f$, we use $\wt{O}(f)$ to denote $f \cdot \poly(\log f)$.

\subsection{Stable Rank and Effective Rank}\label{sec:preli:s_rank_and_e_rank}


\begin{definition}[Stable rank \cite{cnw15}]
\label{def:srank}
Let $A \in \R^{n \times d}$
\begin{align*}
    \mathrm{srank}(A):=\frac{ \| A \|_F^2}{ \| A \|^2}
\end{align*}
to denote the stable rank of $A$.
\end{definition}

\begin{definition}[effective rank]
\label{def:erank}
Let $A \in \R^{d \times d}$, we use
\begin{align*}
    \mathrm{erank}(A):=\frac{ \tr[A]}{ \| A \|}
\end{align*}
to denote the effective rank of $A$.
\end{definition}

\subsection{Basic Algebras}\label{sec:preli:algebra}
\begin{fact}\label{fac:computations}
\begin{itemize}
    \item Let $X \in \R^{k \times k}$, $a \in \R^{k}$, then 
    \begin{align*}
        a^\top X a = \sum_{i=1}^k \sum_{j=1}^k a_i X_{i,j} a_j = \sum_{i=1}^k a_i X_{i,i} a_i + \sum_{i\neq j} a_i X_{i,j} a_j.
    \end{align*}
\end{itemize}
\end{fact}

\subsection{Tools for Matrix Inequality}\label{sec:preli:matrix_norm}
\begin{fact}\label{fac:matrix_bounds}
    Let $A,B \in \R^{n \times d}$, then
    \begin{itemize}
        \item $\|A\|_F \leq \sqrt{\rank(A)} \cdot \|A\|$
        \item $\rank(A + B) \leq \rank(A) + \rank(B)$
        \item $\| A^\top \| = \| A \|$
        \item $\| A \| \geq \| B \| - \| A - B \|$
        \item $\| A + B \| \leq \| A \| + \| B \|$
        \item $\| A \cdot B \| \leq \| A \| \cdot \| B \|$ 
        \item Let $a \in \R$, if $A \preceq a \cdot B$, then $\| A \| \leq a \cdot \| B \|$
        \item Let $a \in \R$, then $\| a \cdot A \| \leq |a| \cdot \| A \|$
        \item Let $x \in \R^d$, we have $\| A x \|_2 \leq \| A \| \cdot \| x \|_2$.
        \item Let $x, y \in \R^d$, then $\| x y^\top \| \leq \| x \|_2 \| y \|_2$
    \end{itemize}
\end{fact}
\subsection{Tools for PSD}\label{sec:preli:psd}

\begin{fact}\label{fac:psd}
    Let $x, y \in \R^d$, We have:
    \begin{itemize}
        \item $xy^\top + y x^\top \preceq x x^\top + y y^\top$
    \end{itemize}
\end{fact}

\begin{fact}\label{fac:more_psd}
Let $\{\alpha_i\}_{i \in[n]} \subseteq \R^d$ be a set of vectors, then we have
\begin{itemize}
    \item Part 1. $a_i a_j^\top + a_j a_i^\top \preceq a_i a_i^\top + a_j a_j^\top$
    \item Part 2. $\sum_{i=1}^n \sum_{j> i}^n a_i a_j^\top + a_j a_i^\top \preceq (n-1) \sum_{i=1}^n a_i a_i^\top$
    \item Part 3. $\sum_{i=1}^n \sum_{j=1}^n a_i a_j^\top \preceq n \cdot \sum_{i=1}^n a_i a_i^\top$
\end{itemize}
\end{fact}
\begin{proof}
{\bf Proof of Part 1}
It trivially follows from Fact~\ref{fac:psd}

{\bf Proof of Part 2.}
We have 
\begin{align*}
\sum_{i=1}^n \sum_{j> i}^n a_i a_j^\top + a_j a_i^\top
\preceq & ~  \sum_{i=1}^n \sum_{j> i}^n (a_i a_i^\top + a_j a_j^\top) \\
= & ~ (n-1) \cdot  \sum_{i=1}^n a_i a_i^\top
\end{align*}
where the first step follows from Part 1. 

{\bf Proof of Part 3.}

\begin{align*}
    \sum_{i=1}^n \sum_{j=1}^n a_i a_j^\top
    = & ~ \sum_{i=1}^n a_i a_i^\top + \sum_{i \neq j} a_i a_j^\top \\
    = & ~ \sum_{i=1}^n a_i a_i^\top + \sum_{i=1}^n \sum_{j> i}^n a_i a_j^\top + a_j a_i^\top \\
    \preceq & ~ n \sum_{i=1}^n a_i a_i^\top,
\end{align*}
where the first step follows from Fact~\ref{fac:psd}, the second step follows from decomposing the second term, and the last step follows from Part 2. 

Thus we complete the proof. 
\end{proof}

\subsection{Basic Definitions}\label{sec:preli:definitions}
\begin{definition}[Regularization Term]\label{def:def_reg}
    Let $A_j \in \R^{n \times d}$, $w \in \R^n$, $W = \diag(w)$. 
We define $L_{\reg} : \R^d \to \R$ as follows
\begin{align*}
L_{\reg,j}(x):= 0.5 \| W A_j x\|_2^2
\end{align*}
\end{definition}
\begin{definition}[Our Softmax Loss Function]
\label{def:softmax_loss}
Let $x \in \R^d$, we define the softmax loss function as follows
\begin{align*}
    L_{\exp}(x):=& \sum_{j=1}^n L_{\exp,j}(x) \\
    L(x) :=& \sum_{j=1}^n L_{\exp,\reg,j}(x)
\end{align*}
where
\begin{align*}
    L_{\exp,j}(x) :=& 0.5 \| \langle \exp(A_j x) , {\bf 1}_n \rangle^{-1} \exp(A_j x) - b_j \|_2^2 \\
    L_{\exp,\reg,j}(x) :=& 0.5 \| \langle \exp(A_j x) , {\bf 1}_n \rangle^{-1} \exp(A_j x) - b_j \|_2^2 + 0.5 \| W A_j x\|_2^2
\end{align*}
$A_j \in \R^{n \times d}$, $b_j \in \R^n$. For a certain batch $\mathcal{B} \in [n]$ of data points, we define
\begin{align*}
    L_{\exp}(x;\mathcal{B}) := \sum_{j \in \mathcal{B}} L_{\exp,j}(x)
\end{align*}
\end{definition}

\begin{lemma}[\cite{dls23}]\label{lem:L_exp_reg_is_strongly_convex}
    Let $L_{\exp,\reg,j}(x) \in \R^d$ follows from Definition~\ref{def:def_reg}, then we have
    \begin{align*}
        \nabla^2 L_{\exp,\reg,j}(x) \succeq \mu \cdot I_d
    \end{align*}
    where $l > 0$ is a constant.
\end{lemma}

\begin{lemma}[Decomposition of gradient, \cite{dls23}]\label{lem:gradient_decomposition}
Given
\begin{itemize}
    \item $L_{\exp,j}(x)$ follows from Definition~\ref{def:softmax_loss}.
    \item $f_j(x)$ follows from Definition~\ref{def:def_f_j}.
    \item $c_j(x)$ follows from Definition~\ref{def:def_c_j}.
\end{itemize}
      Then it holds
    \begin{itemize}
    \item Part 1.
    \begin{align*}
        \nabla L_{\exp, j}(x) = A_j^\top \cdot G_j(x). 
    \end{align*}
    \item Part 2. For ${\cal B} \subseteq [n]$
    \begin{align*}
        \nabla L_{\exp}(x; \mathcal{B}) = \sum_{j \in \mathcal{B}}\nabla L_{\exp, j}(x)
    \end{align*}
    \item Part 3.
    \begin{align*}
        \nabla L_{\exp}(x) = \sum_{j \in [n]}\nabla L_{\exp, j}(x)
    \end{align*}
    \end{itemize}
\end{lemma}

\begin{definition}\label{def:def_f_j}
    We define $f_j(x)$ as follows
    \begin{align*}
        f_j(x) := \langle\exp(A_jx), \mathbf{1}_n\rangle^{-1}\cdot \exp(A_jx).
    \end{align*}
\end{definition}

\begin{definition}\label{def:def_c_j}
Let $b_j \in \R^n$.

We define $c_j(x)$ as 
\begin{align*}
    c_j(x) := f_j(x) - b_j
\end{align*}
\end{definition}

\begin{definition}[\cite{dls23}]\label{def:def_B_j}
    We define $B_j(x)$ as follows
    \begin{align*}
        B_j(x):= & ~ \langle 3 f_j(x) - 2 b_j, f_j(x) \rangle f_j(x) f_j(x)^\top \\
        & ~ + (b_j \circ f_j(x)) f_j(x)^\top + f_j(x) (b_j \circ f_j(x))^\top \\
        & ~ + \langle f_j(x) - b_j, f_j(x) \rangle \cdot \mathrm{diag}( f_j(x) ) \\
        & ~ + \mathrm{diag}( (2f_j(x) - b_j) \circ f_j(x) ) 
\end{align*}
\end{definition}

\begin{definition}\label{def:def_G_j}
Given
\begin{itemize}
    \item Let $f_j(x)$ be defined as Definition~\ref{def:def_f_j}.
    \item Let $c_j(x)$ be defined as Definition~\ref{def:def_c_j}.
\end{itemize}
    We define $G_j : \R^d \rightarrow \R^k$ as follows
    \begin{align*}
        G_j(x) :=  - \underbrace{ f_j(x) }_{k \times 1} \underbrace{ c_j(x)^\top }_{1 \times k} \underbrace{ f_j(x) }_{k \times 1} + \underbrace{ \diag(f_j(x)) }_{k \times k} \underbrace{ c_j(x) }_{k \times 1} 
    \end{align*}
\end{definition}

For convenient, we define
\begin{definition}\label{def:def_G_j_1_2}
Given
\begin{itemize}
    \item $f_j(x)$ follows from Definition~\ref{def:def_f_j}.
    \item $b_j \in \R^k$
\end{itemize}
We define $G_{j,1} : \R^d \rightarrow \R^k$ and $G_{j,2} : \R^{d} \rightarrow \R^k$
\begin{itemize}
    \item $G_{j,1}:= f_j(x)(f_j(x) - b_j)^\top f_j(x)$
    \item $G_{j,2}:= \mathrm{diag}(f_j(x))(f_j(x) - b_j)$
\end{itemize}
Then it is obvious that $G_j(x) = -G_{j,1}(x) + G_{j,2}(x)$ (see Definition~\ref{def:def_G_j}).
\end{definition}

\begin{lemma}
\label{lem:decompose_Hessian}
Let $B_j(x)$ be defined as Definition~\ref{def:def_B_j} and $L_{\exp}(x)$ be defined as Definition~\ref{def:softmax_loss}, then we have
\begin{align*}
    \nabla^2 L_{\exp}(x) = \sum_{j=1}^n A_j^\top B_j(x) A_j
\end{align*}
\end{lemma}
\begin{proof}
    It trivially follows from Lemma 5.10 of \cite{dls23}.
\end{proof}

\subsection{Definition of General Properties}\label{sec:general_defs}

\begin{definition}[$l$-Smooth]
\label{def:l_smooth}
    We say a differentiable function $L(x): \R^d \rightarrow \R$ is $l$-smooth if 
    \begin{align*}
        \|\nabla L(x) - \nabla L(y)\|_2 \le l \cdot \|x-y\|_2, ~~\forall x, y \in \R^d.
    \end{align*}
\end{definition}

\begin{definition}[Stong Convexity]
\label{def:strong_convex}
    We say a continuously differentiable function $f:\R^d \rightarrow \R$ is strongly convex if there exists a possitive number $\mu$ such that
    \begin{align*}
        f(y) \ge f(x) \nabla f(x)^\top (y-x) + \frac{1}{2} \mu \|y-x\|_2^2, ~~\forall x, y \in\R^d. 
    \end{align*}
    Equivalently, if the function is twice differentiable, then
    \begin{align*}
        f(x)\text{ is } \mu-\text{strongly convex}
        \iff \nabla^2 f(x) \succeq \mu I. 
    \end{align*}
\end{definition}

\begin{definition}[Polyak-{\L}ojasiewicz Inequality]
\label{def:pl_inequality}
    We say a function $L(x) : \R^d \rightarrow \R$ satisfies $\mu$-Polyak-{\L}ojasiewicz (PL) inequality if for all $x \in \R^d$, it holds that
    \begin{align*}
        \frac{1}{2}\|\nabla L(x)\|^2 \ge \mu(L(x) - L^*),
    \end{align*}
    where $L^* := \min_{x \in \R^d} L(x)$. 
\end{definition}

We have the following existing lemma connecting strong-convexity and PL inequality. 

\begin{lemma}[\cite{kns16}]
\label{lem:pl_conv}
    If a function $L(x)$ is $\mu$-strongly convex, then it is $\mu$-PL. 
\end{lemma}

\subsection{Simultaneous Perturbation Stochastic Approximation (SPSA)}\label{sec:preli:spsa}
\begin{definition}[Simultaneous Perturbation Stochastic Approximation (SPSA) \cite{Spa92}]
\label{def:spsa}
    Let $L(x)$ be a loss function. For a point $x_0 \in \R^d$, we define the Simultaneous Perturbation Stochastic Approximation (SPSA) of $L(x)$ on $x_0$ as a vector $\wh{g}(x_0) \in \R^d$ such that
    \begin{align*}
        \wh{g}(x_0) := \frac{L(x_0 + \epsilon\cdot p) - L(x_0 - \epsilon\cdot p)}{2\epsilon}\cdot p, ~~\forall i \in [d],
    \end{align*}
    where $p \in \R^d \sim \mathcal{N}(0, I_d)$ is the perturbation vector and $\epsilon > 0$ is the perturbation scale. 
\end{definition}

\begin{remark}[$k$-SPSA]
\label{remark:k_spsa}
    The $k$-SPSA gradient estimate averages $\hat{g}(x)$ over $k$ randomly sampled $z$.
\end{remark}

\begin{lemma}[\cite{Spa92}]
    The gradient estimate $\wh{g}(x)$ is almost unbiased, i.e.,
    \begin{align*}
        \E[\wh{g}(x)|x] = pp^\top\nabla L(x),
    \end{align*}
    with probability of $1$. 
\end{lemma}

\subsection{Previous Results}\label{sec:preli:previous}

\begin{lemma}[Lemma~2 in \cite{mgn+23}]
\label{lem:gradient_expect}
    Let $L$ be defined as Definition~\ref{def:softmax_loss}, then we have 
    \begin{align*}
        \E[\|\wh{g}(x, \mathcal{B})\|^2] = \frac{d + k - 1}{k}\cdot \E[\|\nabla L(x, \mathcal{B})\|^2],
    \end{align*}
    where $k$ is the parameter for $k$-SPSA. 
\end{lemma}

\begin{definition}[Gradient Covariance]
\label{def:grad_covv}
    We say the covariance of SGD gradient estimate on a minibatch $\mathcal{B}$ of size $B$ is defined as
    \begin{align*}
        \Sigma(x) = B(\E[\nabla L(x;\mathcal{B})\nabla L(x; \mathcal{B})^\top] - \nabla L(x;\mathcal{B})\nabla L(x; \mathcal{B})^\top).
    \end{align*}
\end{definition}

\begin{lemma}[Lemma~5 in \cite{mgn+23}]
\label{lem:prop_gauss}
    Let $z \in \R^n$ with $z_i \sim \N(0, 1)$ i.i.d. Then it holds that
    \begin{align*}
       &~ \E[\wh{g}(x, \mathcal{B})\wh{g}(x, \mathcal{B})^\top] \\
    =  &~ (1+\frac{1}{n})\cdot (\nabla L(x)L(x)^\top +\frac{1}{B}\Sigma(x)) + \frac{1}{n}I\cdot (\|\nabla L(x)\|^2 + \frac{1}{B}\tr(\Sigma(x))). 
    \end{align*}
\end{lemma}

\section{Analysis for Softmax Function}

In this section, we provide analysis for the softmax loss function. 
In Section~\ref{sec:softmax_smooth} we proved that the softmax loss function is smooth.
In Section~\ref{sec:previous_work_tools} we state some useful lemmas from our previous work.
In Section~\ref{sec:G_j_1_is_smooth} we proved that $G_{j,1}$ is smooth.
In Section~\ref{sec:G_j_2_is_smooth} we proved that $G_{j,2}$ is smooth.
In Section~\ref{sec:effective_bound} we find the upper bound of the effecive rank of $H$ by upper bounding the stable rank of $B(x)$.
In Section~\ref{sec:connections} we state the inequality between stable rank and effective rank.

\subsection{Softmax Loss is Smooth}
\label{sec:softmax_smooth}
We have the following lemma
\begin{lemma}
\label{lem:softmax_smmoth}
    Given
    \begin{itemize} 
        \item $A \in \R^{n \times d}$
        \item $R \geq 4$
        \item $x, y \in \R^d$ satisfy $\| A (x-y) \|_{\infty} < 0.01$
        \item $\| A \| \leq R$
        \item Let $R_f:= n^{1.5} \exp(5 R^2)$ 
        \item Let $W = \diag(w)$, where $w_i^2 \le \frac{1}{\sigma_{\max}(A)}$. 
    \end{itemize}
    Then the Softmax loss function (Definition~\ref{def:softmax_loss}) $L_{j}(x)$ is $l$-smooth (Definition~\ref{def:l_smooth}), where
    \begin{align*}
        l = 8 R R_f.
    \end{align*}
\end{lemma}

\begin{proof}
    Let $x, y \in \R^d$ be two arbitrary point. By Lemma~\ref{lem:first_smooth} and Lemma~\ref{lem:second_smooth} we have
    \begin{align*}
        &~\|\nabla L_{\exp, j}(x) - \nabla L_{\exp, j}(y)\|_2 \\
    \le &~\|A_j\|_2 \cdot (\|G_1(x) - G_1(y)\|_2 + \|G_2(x) - G_2(y)\|_2) \\
    \le &~ 8RR_f \cdot \|x - y\|_2 
    \end{align*}
    where the first step follows from definition of $G_1$ and $G_2$, the second step follows from Fact~\ref{fac:known_bounds}, Lemma~\ref{lem:first_smooth} and Lemma~\ref{lem:second_smooth}. 

    Trivially,
    \begin{align*}
        \nabla L_{\mathrm{reg},j}(x) = AW^2A^\top x.
    \end{align*}
    Thus we have
    \begin{align*}
        &~\|\nabla L_{\mathrm{reg}, j}(x) - \nabla L_{\mathrm{reg}, j}(y)\|_2 \\
    \le &~\|AW^2A^\top\|_2 \cdot \|x-y\|_2 \\
    \le &~ \|x - y\|_2, 
    \end{align*}
    where the first step follows from definition of spectral norm, the second step follows from $w_i^2 \le \frac{1}{\sigma_{\max}(A)}$. 

    Adding the above together, we have $l = 8RR_f + 1$. Since $8RR_f\gg 1$ trivially, we complete the proof. 
\end{proof}

\subsection{Tools from previous work}\label{sec:previous_work_tools}


\begin{lemma}[Lemma~5.2 in \cite{dls23}]
\label{lem:basic_f}
Let $f_j : \R^d \rightarrow \R$ 
follows from Definition~\ref{def:def_f_j}, then for $\forall x \in \R^d$, it holds  
\begin{itemize}
    \item $\| f_j(x) \|_2 \leq \| f_j(x) \|_1 \leq 1$.
    \item $0 \preceq f_j(x) f_j(x)^\top \preceq I_n$.
    \item Let $b \in \R^d$, $0 \preceq (b \circ f_j(x)) (b\circ f_j(x))^\top \preceq \| b \|_{\infty}^2 f_j(x) f_j(x)^\top \preceq \| b \|_{\infty}^2 I_n$
    \item Let $b \in \R^d$, $\mathrm{diag} (b \circ b) \preceq \| b \|_{\infty}^2 I_n$
    \item $0 \preceq \mathrm{diag}(f_j(x)) \preceq \| f_j(x) \|_{\infty} I_n \preceq  \| f_j(x) \|_2 I_n$.
    \item $0 \preceq \mathrm{diag}(f_j(x) \circ f_j(x)) \preceq \| f_j(x) \|_{\infty}^2 I_n \preceq \| f_j(x) \|_2 I_n$.
\end{itemize}
\end{lemma}

\begin{fact}[Lemma~7.2 in \cite{dls23}]\label{fac:dls23_parameters}
\label{fac:known_bounds}
    If the following conditions hold 
    \begin{itemize} 
        \item Let $A \in \R^{n \times d}$
        \item Let $R \geq 4$
        \item Let $x, y \in \R^d$ satisfy $\| A (x-y) \|_{\infty} < 0.01$
        \item $\| A \| \leq R$
        \item Let $R_f:= n^{1.5} \exp(5 R^2)$ 
    \end{itemize}
    We have
    \begin{itemize}
        \item Part 0. $\| \exp(Ax) \|_2 \leq \sqrt{n} \exp(R^2)$
        \item Part 1. $\| \exp(Ax) - \exp(Ay) \|_2 \leq 2 \sqrt{n} R \exp(R^2) \cdot \| x - y \|_2$
        \item Part 2. $\| f_j(x) - f(y) \|_2 \leq R_f \cdot \| x - y \|_2$
    \end{itemize}
\end{fact}

\begin{lemma}[\cite{dls23}]\label{lem:beta}
If the following conditions holds
\begin{itemize}
    \item $\| A \| \leq R$
    \item $\| x \|_2 \leq R$
    \item Let $\beta$ be lower bound on $\langle \exp(Ax), {\bf 1}_n \rangle$
\end{itemize}
Then we have
\begin{align*}
    \beta \geq \exp(-R^2)
\end{align*}
\end{lemma}

\subsection{Smoothness for function \texorpdfstring{$G_{j,1}$}{}}\label{sec:G_j_1_is_smooth}

\begin{lemma}
\label{lem:first_smooth}
    We define 
    \begin{align*}
        G_{j,1}(x) := f_j(x)(f_j(x) - b_j)^\top f_j(x). 
    \end{align*}
    Then we have 
    \begin{align*}
        \|G_{j,1}(x) - G_{j,1}(y)\|_2 \le 5R_f \cdot \|x - y\|_2. 
    \end{align*}
\end{lemma}

\begin{proof}
    Since $f_j(x), b_j \in \R$, thus we have
    \begin{align*}
        G_1(x) = (f_j(x))^3 - (f_j(x))^2 b_j.
    \end{align*}
    Then by Fact~\ref{fac:known_bounds}, we have
    \begin{align*}
        &~\|G_1(x) - G_1(y)\|_2 \\
    \le &~\|(f_j(x))^3 - (f_j(y))^3\|_2 + \|(f_j(x))^2 - (f_j(y))^2\|_2 \\
    \le &~5\|f_j(x) - f_j(y)\|_2 \\
    \le &~ 5 R_f \|x - y\|_2
    \end{align*}
    where the first step follows from triangle inequality and Fact~\ref{fac:known_bounds}, the second step follows from Lemma~\ref{lem:basic_f}, the last step follows from Fact~\ref{fac:known_bounds}. 
\end{proof}

\subsection{Smoothness for function \texorpdfstring{$G_{j,2}$}{}}\label{sec:G_j_2_is_smooth}

\begin{lemma}
\label{lem:second_smooth}
    We define 
    \begin{align*}
        G_{j,2}(x) := \mathrm{diag}(f_j(x))(f_j(x) - b_j).
    \end{align*}
    Then we have
    \begin{align*}
        \|G_{j,2} (x) - G_{j,2}(y)\|_2 \le 3 R_f \cdot \|x - y\|_2.
    \end{align*}
\end{lemma}

\begin{proof}
    We have
    \begin{align*}
        &~\|G_2(x) - G_2(y)\|_2 \\
    \le &~\|(f_j(x))^2 - (f_j(y))^2\|_2 + \|f_j(x) - f_j(y)\|_2 \\
    \le &~ 3\|f_j(x) - f_j(y)\|_2 \\
    \le &~ 3R_f\|x-y\|_2
    \end{align*}
    where the first step follows from $f_j(x) \in \R$ and Fact~\ref{fac:known_bounds}, the second step follows from Lemma~\ref{lem:basic_f}, the last step follows from Fact~\ref{fac:known_bounds}. 
\end{proof}

\subsection{Effective Bound for \texorpdfstring{$H$}{}}\label{sec:effective_bound}



\begin{lemma}[Upper Bound Stable Rank of $B_j$]
\label{lem:upper_srank_B}
    Let $B_j$ be defined as in Lemma~\ref{lem:decompose_Hessian}, then we have
    \begin{align*}
        \mathrm{srank}(B_j) = (\frac{\|B_j\|_F}{\|B_j\|})^2 \le 2d + 2,
    \end{align*}
    where $\mathrm{srank}$ is defined as Definition~\ref{def:srank}. 
\end{lemma}

\begin{proof}
    Firstly, we have
    \begin{align*}
        \frac{\|B_j\|_F}{\|B_j\|}
        \le & ~ \frac{\sqrt{\rank(B_j)}\|B_j\|}{\|B_j\|} \\
        = & ~ \sqrt{\rank(B_j)}
    \end{align*}
    where the first step follows from Fact~\ref{fac:matrix_bounds},
    the second step follows from simple algebra.

    Secondly, by applying Lemma 5.15 of \cite{dls23}, we can show that $B_j$ is composed of several rank-$1$ matrices and diagonal matrices:
    \begin{align*}
        B_{\rank,j}(x) := & ~ \underbrace{ \langle 3 f_j(x) - 2 b_j, f_j(x) \rangle f_j(x) f_j(x)^\top  }_{ :=B_{\rank,j}^1(x) } + \underbrace{ (b_j \circ f_j(x)) f_j(x)^\top + f_j(x) (b_j \circ f_j(x))^\top }_{ :=B_{\rank,j}^2(x) } \\
    B_{\diag,j}(x):= & ~ \underbrace{ \langle f_j(x) - b_j, f_j(x) \rangle \cdot \diag( f_j(x) ) }_{ := B_{\diag,j}^1(x)}  + \underbrace{ \diag( (2f_j(x) - b_j) \circ f_j(x) ) }_{ := B_{\diag,j}^2(x) }
    \end{align*}

    Thus, we can bound $\rank(B_j)$ as follows
    \begin{align*}
        \rank(B_j)
        = & ~ \rank(B_{\rank,j} + B_{\diag,j}) \\
        \leq & ~ \rank(B_{\rank,j}) + \rank(B_{\diag,j}) \\
        = & ~ \rank(B_{\rank,j}^1 + B_{\rank,j}^2) + \rank(B_{\diag,j}^1 + B_{\diag,j}^2) \\
        \leq & ~ \rank(B_{\rank,j}^1) + \rank(B_{\rank,j}^2) + \rank(B_{\diag,j}^1 ) + \rank(B_{\diag,j}^2) \\
        \leq & ~ 1 + 1 + d + d \\
        = & ~ 2d + 2
    \end{align*}
    where the first step follows from decomposing $B_j$,
    the second step follows from Fact~\ref{fac:matrix_bounds},
    the third step follows from decomposing $B_{\rank,j},B_{\diag,j}$,
    the fifth step follows from $\rank(B_{\rank,j}^1) = \rank(B_{\rank,j}^2) = 1$ and $\rank(B_{\diag,*}) = \nnz(B_{\diag,*}) \leq d$ for $B_{\diag,*} \in \R^{d \times d}$,
    the last step follows from simple algebra.

    Thus, we aquired the bound for $(\frac{\|B_j\|_F}{\|B_j\|})^2$:
    \begin{align*}
        \frac{\|B_j\|_F}{\|B_j\|} \leq \sqrt{2d + 2}
        \Longrightarrow (\frac{\|B_j\|_F}{\|B_j\|})^2 \leq 2d + 2
    \end{align*}

\end{proof}

\subsection{The connection between effective rank and stable rank}\label{sec:connections}
The following lemma provide upper bound for the effective rank of $H$, in the term of stable rank of $B$. 
\begin{lemma}
\label{lem:e_rank_H}
Let $A \in \R^{n \times d}, B \in \R^{n \times n}$ be two matrix. If the following conditions hold
\begin{itemize}
    \item $\|B\|_F/\|B\| \le r$ 
    \item Let $H = A^\top B A$
\end{itemize}
Then,
\begin{align*}
     \mathrm{erank}(H) \leq \rank(A)\cdot r \cdot \kappa^2(A),
\end{align*}
where $\mathrm{erank}$ is defined as Definition~\ref{def:erank}. 
Without loss of generality, we can assume $n \gg d$, then
\begin{align*}
    \mathrm{erank}(H) \leq d r \cdot \kappa^2(A).
\end{align*}
\end{lemma}

\begin{proof}
    We have
    \begin{align}
    \label{eq:upper_trace}
        \tr[H] 
    = &~\tr[A^\top B A] \notag\\
    = &~\tr[A A^\top B] \notag\\
    \le &~\|AA^\top\|_F \cdot \|B\|_F \notag\\
    \le &~\|A\|_F^2 \cdot \|B\|_F \notag \\
    \le &~ \rank(A) \cdot \sigma_{\max}^2(A) \cdot \|B\|_F,
    \end{align}
    where the first step follows from definition of $H$, the second step follows from the cyclic rule of matrix trace, the third and fourth steps follows from the Cauchy-Schwartz inequality, and the last step follows from the definition of Frobenius norm. 

    Now we provide the lower bound for the spectral norm of $\|H\|$. We have
    \begin{align}
    \label{eq:lower_norm}
        \|H\| 
    = \|A^\top B A\|
    \le \sigma_{\min}^2(A) \cdot \|B\|. 
    \end{align}
    Thus by Eq.~\eqref{eq:upper_trace} and Eq.~\eqref{eq:lower_norm} we have,
    \begin{align*}
        \tr[H]/\|H\|
    \le \frac{\rank(A) \cdot \sigma_{\max}^2(A) \cdot \|B\|_F}{\sigma_{\min}^2(A) \cdot \|B\|} \le \rank(A) \cdot r \cdot \kappa^2(A).
    \end{align*}
    Thus we completed the proof. 
\end{proof}     

\section{Loss Analysis for Gradient Descent}

Here in this section, we provide analysis for the loss in each iteration of the Gradient Descent. 
In Section~\ref{sec:def_gradient_step}, we define how we update the parameters in traditional SGD.
In Sextion~\ref{sec:loss_decrease}, we analyze the decrease of loss per iteration.

\subsection{Gradient Step}\label{sec:def_gradient_step}

\begin{definition}[GD step]
\label{def:gd_step}
    The gradient descent step based on the zero-th order method is defined as
    \begin{align*}
        x_{t+1} \gets x_t - \eta \cdot \wh{g}(x_t),
    \end{align*}
    where $\wh{g}(x_t)$ is defined as Definition~\ref{def:spsa}. 
\end{definition}


\subsection{Loss Decrease}\label{sec:loss_decrease}

We have the following convergence lemma. 

\begin{lemma}[Convergence Rate
]
\label{lem:converge_rate}
    Let $x_{t+1} \gets x_t - \eta \wh{g}(x_t)$, where $\wh{g}(x_t)$ is computed with respect to the batch $\mathcal{B}$. Consider $L_{\exp}(x)$ as defined in Definition~\ref{def:softmax_loss}, then there exists a parameter 
    \begin{align*}
        \gamma =
    \frac{d^2 \cdot \sqrt{2d+2} \cdot \kappa^2(A) + d - 2}{k(d + 2)} + 1
    \end{align*}
    such that the expected loss decrease can be bounded as 
    \begin{align*}
        \E[L(x_{t+1} ) |x_t] - L(x_t) \le -\eta \|\nabla L(x_t)\|^2 + \frac{1}{2} \eta^2 \ell \cdot \gamma \cdot \E[\|\nabla L(x; \mathcal{B})\|^2]
    \end{align*}
\end{lemma}

\begin{proof}
    By Taylor’s theorem with remainder, we have that
    \begin{align}
    \label{eq:loss_taylor}
        L(x_{t+1})
    = &~L(x_{t}) +\nabla L(x_t)^\top (x_{t+1} - x_t) \notag\\
      &~ + \int_{0}^1 \lambda(x_{t+1} - x_t)^\top \nabla^2L(\lambda x_{t+1} + (1 - \lambda)x_t)(x_{t+1} - x)^\top d\lambda.
    \end{align}
    
    Then by 
    \begin{align*}
        \|x_{t+1} - x_t\| 
    = &~\eta \cdot \|\wh{g}(x; \mathcal{B})\| \\
    \le&~\eta \sqrt{d} \cdot \frac{1}{kB}\sum_{i = 1}^k\sum_{j = 1}^B |z_i^\top \nabla L_{\exp,j}(x)| \\
    \le&~\eta d G_{\max}(x_t),
    \end{align*}
    where $G_{\max}:=\max_{j \in [n]}\nabla L_{\exp,j}(x_t)$.
    The first step follows from the definition of GD step,
    the second step follows from the way we calculate $\wh{g}$ ($k$-SPSA in Remark~\ref{remark:k_spsa})
    , the third step follows from $|z_i^\top \nabla L_{\exp,j}(x)| \leq G_{\max}(x_t)$ and $\sqrt{d} \leq d$.
    
    Thus we have
    \begin{align}
    \label{eq:interpolation}
        \|\lambda x_{t+1} +(1 - \lambda)x_t - x_t\| \le \eta d G_{\max}(x_t).
    \end{align}
    this follows from simple algebra.
    
    
    We define
    \begin{align}
    \label{eq:def_H_lambda}
        H_{\lambda}(x_t) := \nabla^2 L(\lambda x_{t+1} +(1 - \lambda)x_t).
    \end{align}
    Then we have
    \begin{align*}
        L(x_{t+1}) 
    \le &~ L(x_t) +\nabla L(x_t)^\top (x_{t+1} - x_t) + (x_{t+1} - x_t)^\top H_{\lambda}(x_t)(x_{t+1} - x_t) \\
    = &~ L(x_t) - \eta \nabla L(x_t)^\top \wh{g}(x_t;\mathcal{B}) + \frac{1}{2}\eta^2 \wh{g}(x_t;\mathcal{B})^\top H_{\lambda}(x_t)\wh{g}(x_t; \mathcal{B}).
    \end{align*}
    where step 1 follows from Eqs.\eqref{eq:loss_taylor}, \eqref{eq:interpolation} and \eqref{eq:def_H_lambda},
    step 2 follows from the way we update $x_t$.

    We have
    \begin{align*}
        \E[L(x_{t+1})|x_t]
    \le &~ L(x_t) - \eta \|\nabla L(x_t)\|^2 +\frac{\eta^2}{2} \langle H_{\lambda}(x_t), \E[\wh{g}(x;\mathcal{B})\wh{g}(x;\mathcal{B})^\top] \rangle \\
    = &~ L(x_t) - \eta \|\nabla L(x_t)\|^2 + \frac{\eta^2}{2} \cdot \frac{d}{k(k(d+2))} \cdot (\|\nabla L(x_t)\|^2 + \frac{1}{B} \tr[\Sigma(x_t))]\tr[H_{\lambda}(x_t)] \\
    &~+ \frac{\eta^2}{2}(1 + \frac{d-2}{k(d+2)})(\nabla L(x_t)^\top H_{\lambda}(x_t)\nabla L(x_t) + \frac{1}{B}\langle \Sigma(x_t), H_{\lambda}(x_t)\rangle).
    \end{align*}
    where step 1 follows from 
    taking conditional expectation with respect to $x_t$, step 2 follows from Lemma~\ref{lem:prop_gauss}.
    
    We have
    \begin{itemize}
        \item Part 1. $\|H_{\lambda}(x_t)\| \le 8RR_f = l$, by Lemma~\ref{lem:softmax_smmoth};
        \item Part 2. $\tr [ H_{\lambda}(x_t) ]/\|H_{\lambda}(x_t)\| \le d \cdot \sqrt{2d+2} \cdot \kappa^2(A) = r$, by Lemma~\ref{lem:upper_srank_B} and Lemma~\ref{lem:e_rank_H}.
    \end{itemize}
    Thus
    \begin{align}\label{eq:upper_boundz_of_tr_h_lambda_x_t}
        \tr[H_{\lambda}(x_t)] \le 8 RR_f \cdot d \cdot \sqrt{2d+2} \cdot \kappa^2(A) = lr.
    \end{align}
    This follows from combining {\bf Part 1} and {\bf Part 2}.
    
    Then we have
    \begin{align*}
        \E[L(x_{t+1})|x_t]
    \le &~ L(x_t) - \eta \|\nabla L(x_t)\|^2 + \frac{\eta^2 l}{2} \cdot (\frac{dr+d-2}{k(d+2)} + 1)\cdot (\|\nabla L(x_t)\|^2 + \frac{1}{B} \tr[\Sigma(x_t)]) \\
    = &~ L(x_t) - \eta \|\nabla L(x_t)\|^2 + \frac{\eta^2 l}{2} \cdot (\frac{dr+d-2}{k(d+2)} + 1) \cdot \E[\|\nabla L(x_t; \mathcal{B})\|^2].
    \end{align*}
    where step 1 follows from Eq.~\eqref{eq:upper_boundz_of_tr_h_lambda_x_t},
    step 2 follows from definition of $\Sigma(x)$ (Definition~\ref{def:grad_covv}).
    
    Defining
    \begin{align*}
        \gamma 
    = &~ \frac{dr+d-2}{k(d+2)} + 1 \\
    = &~ \frac{d^2 \cdot \sqrt{2d+2} \cdot \kappa^2(A) + d - 2}{k(d + 2)} + 1. 
    \end{align*}
    and we complete the proof.
\end{proof}

We also have the following result. 

\begin{corollary}
\label{cor:zo_and_sgd}
    By Lemma~\ref{lem:converge_rate} and Lemma~\ref{lem:gradient_expect}, we choose $\eta = \eta_0$, where $\eta_0$ is used in traditional SGD. Then we have
    \begin{align*}
        \E[L(x_{t+1} ) |x_t] - L(x_t) \le \frac{1}{\gamma} \cdot ( -\eta_0 \|\nabla L(x_t)\|^2 + \frac{1}{2} \eta_0^2 \ell \cdot \E[\|\nabla L(x; \mathcal{B})\|^2] ). 
    \end{align*}
\end{corollary}
\section{Convergence Analysis}

In this section, we provide the analysis for convergence of our algorithm. During this section, we use $L^*:=\min_{x \in \R^d} L(x)$ to denote the global minimum of $L(x)$. 
In Section~\ref{sec:softmax_strongly_convex}, we proved that $L_{\exp,\reg}$ is strongly convex and thus is PL.
In Section~\ref{sec:upper_bound_cov}, we upper bound the trace of covariance matrix under certain assumptions.
In Section~\ref{sec:sgd_previous}, we state an existing result with respect to the traditional SGD.
In Section~\ref{sec:global_convergence}, we provide our main result, we show that our algorithm has convergence guarantee for softmax loss function.

\subsection{Softmax Loss is Strongly Convex}\label{sec:softmax_strongly_convex}
We have the following lemma
\begin{lemma}
\label{lem:softmax_convex}
    Let $L_{\exp,\reg, j}(x)$ be defined as Definition~\ref{def:softmax_loss}, then there exists a parameter $\mu$ such that it is $\mu$-strongly convex (Definition~\ref{def:strong_convex}). 
    And by Lemma~\ref{lem:pl_conv}, it is also $\mu$-PL. 
\end{lemma}

\begin{proof}
    By the definition of strongly convex, we know that if a function $f(x)$ is strongly convex, then
    \begin{align*}
        \nabla^2 f(x) \succeq \mu I
    \end{align*}
    where $\mu$ is a positive constant.

    Thus, by applying Lemma~\ref{lem:L_exp_reg_is_strongly_convex}, $L_{\exp,\reg,j}$ is strongly convex with parameter $\mu$. 
\end{proof}

\subsection{Upper Bound Covariance}\label{sec:upper_bound_cov}
\begin{lemma}
\label{lem:softmax_upper_bound_covv}
    Let $\Sigma(x)$ be defined as Definition~\ref{def:grad_covv},
    If 
    \begin{align}
    \label{eq:upper_alpha_assump}
        \tr[ \sum_{j\in [n]} A_j^\top G_j(x) G_j(x)^\top A_j ] \preceq \epsilon_0^{-1} \frac{n}{B^2} \alpha ( L(x) - L^* )
    \end{align}
    
    Then we have 
    \begin{align*}
        \tr[\Sigma(x)] \le  \alpha \cdot (L(x) - L^*),
    \end{align*}
    for all $x \in \R^d$. 
\end{lemma}

\begin{proof}

We have
\begin{align*}
    \tr[ \Sigma(x) ] 
    \leq & ~ | \tr[\Sigma(x)] | \\
    = & ~ | \tr[ \nabla L (x; {\cal B} ) \nabla L(x ; {\cal B})^\top  - \E[  \nabla L (x; {\cal B} ) \nabla L(x ; {\cal B})^\top  ] ] | \\
    \leq & ~ \epsilon_0 \cdot \frac{B^2}{n} \tr[ \sum_{j \in [n]} A_j^\top G_j(x) G_j(x)^\top A_j ] \\
    \leq & ~ \alpha \cdot (L(x) - L^*)
\end{align*}
where step 1 follows from simple algebra, step 2 follows from the definition of $\Sigma(x)$, 
step 3 follows from Assumption~\ref{ass:approximation} and Lemma~\ref{lem:rewrite_rank_1_gradient}, step 4 follows from Eq.\eqref{eq:upper_alpha_assump}. 
\end{proof}

\begin{assumption}\label{ass:approximation}
Let $\epsilon_0 = 1/4$. 
We assume the following balanced distribution, for all ${\cal B} \subset [n]$
\begin{itemize} 
\item $\sum_{j \in {\cal B}} A_j^\top G_j(x) G_j(x)^\top A_j \approx ( 1 \pm \epsilon_0 ) \frac{B}{n} \sum_{j \in [n]} A_j^\top G_j(x) G_j(x)^\top A_j $ 
\item $\sum_{j_1 \neq j_2 \in {\cal B}} A_{j_1}^\top G_{j_1}(x) G_{j_2}(x)^\top A_{j_2} \approx (1 \pm \epsilon_0 ) B(B-1) \cdot \frac{1}{n} ( \sum_{j\in[n]} A_j^\top G_j(x) G_j(x)^\top A_j )$
\end{itemize}
\end{assumption}

\begin{lemma}\label{lem:rewrite_rank_1_gradient}
Given
\begin{itemize}
    \item $L (x; {\cal B} )$ follows from Definition~\ref{def:softmax_loss}
    \item $G_j(x)$ follows from Definition~\ref{def:def_G_j}, for$\forall j \in [n]$
\end{itemize}
Then we can show
\begin{itemize}
    \item Part 1.
    \begin{align*}
        \nabla L (x; {\cal B} ) \nabla L(x ; {\cal B})^\top = & ~ \sum_{j\in {\cal B}} A_j^\top G_j(x) G_j(x) A_j  + \sum_{j_1 \neq j_2 \in {\cal B}} A_{j_1}^\top G_{j_1}(x) G_{j_2}(x) A_{j_2}
    \end{align*}
    \item Part 2.
    \begin{align*}
        \E[ \sum_{j\in {\cal B}} A_j^\top G_j(x) G_j(x) A_j ] = \frac{B}{n} \sum_{j \in [n]} A_j^\top G_j(x) G_j(x) A_j
    \end{align*}
    \item Part 3.
    \begin{align*}
        \E[ \sum_{j_1 \neq j_2 \in {\cal B}} A_{j_1}^\top G_{j_1}(x) G_{j_2}(x) A_{j_2} ] = & ~ B(B-1) \cdot ( \frac{1}{n} \sum_{j\in[n]} A_j^\top G_j(x) ) ( \frac{1}{n} \sum_{j\in [n]} G_j(x)^\top A_j ) \\
        \preceq & ~ B(B-1) \frac{1}{n} \sum_{j \in [n]} A_j^\top G_j(x) G_j(x) A_j
    \end{align*}
\end{itemize}
\end{lemma}
\begin{proof}

{\bf Proof of Part 1.}
By applying Lemma~\ref{lem:gradient_decomposition}, we have
\begin{align*}
    \nabla L (x; {\cal B} ) \nabla L(x ; {\cal B})^\top
    = & ~ \sum_{j_1 \in \mathcal{B}}\nabla A_{j_1}^\top G_{j_1}(x) (\sum_{j_2 \in \mathcal{B}}\nabla A_{j_2}^\top G_{j_2}(x))^\top \\
    = & ~ \sum_{j_1,j_2 \in \mathcal{B}} A_{j_1}^\top G_{j_1}(x) G_{j_2}(x)^\top A_{j_2}
\end{align*}
where step 1 follows from Lemma~\ref{lem:gradient_decomposition},
step 2 follows from simple algebra.

Then, by applying Fact~\ref{fac:computations} we have
\begin{align*}
    \sum_{j_1,j_2 \in \mathcal{B}} A_{j_1}^\top G_{j_1}(x) G_{j_2}(x)^\top A_{j_2}
    = & ~ \sum_{j\in {\cal B}} A_j^\top G_j(x) G_j(x)^\top A_j  + \sum_{j_1 \neq j_2 \in {\cal B}} A_{j_1}^\top G_{j_1}(x) G_{j_2}(x)^\top A_{j_2}
\end{align*}

Thus, we completes the proof.

{\bf Proof of Part 2}
We have
\begin{align*}
    \E[ \sum_{j\in {\cal B}} A_j^\top G_j(x) G_j(x) A_j ]
    = & ~ \frac{B}{n} \sum_{j \in [n]} A_j^\top G_j(x) G_j(x) A_j
\end{align*}
this follows from expectation.

{\bf Proof of Part 3}
We have
\begin{align*}
    \E[\sum_{j_1 \neq j_2 \in {\cal B}} A_{j_1}^\top G_{j_1}(x) G_{j_2}(x) A_{j_2}]
    = & ~  B(B-1) \cdot ( \frac{1}{n} \sum_{j\in[n]} A_j^\top G_j(x) ) ( \frac{1}{n} \sum_{j\in [n]} G_j(x)^\top A_j ) \\
    \preceq  & ~ B(B-1) \frac{1}{n} \sum_{j \in [n]} A_j^\top G_j(x) G_j(x) A_j
\end{align*}
where step 1 follows from expectation, step 2 follows from Fact~\ref{fac:more_psd}.

\end{proof}

\subsection{Previous Results on SGD}\label{sec:sgd_previous} 
\begin{lemma}[Lemma~4 in \cite{mgn+23}]
\label{lem:sgd_conv}
    Assume a loss function satisfies 
    \begin{itemize}
        \item $\mu$-PL (Definition~\ref{def:pl_inequality});
        \item It holds that $
                \tr[\Sigma(x)] \le \alpha \cdot (L(x) - L^*)$;
        \item $l$-smooth;
        \item Its Hessian $H$ satisfies $\mathrm{erank}(H) \le r$. 
    \end{itemize}
    Then after 
    \begin{align*}
        O((\frac{l}{\mu} + \frac{l\alpha}{\mu^2 B})\cdot \log\frac{L(x_0) - L^*}{\epsilon}))
    \end{align*}
    iterations of SGD (with the real gradient), it holds that
    \begin{align*}
        \E[L(x_t)] \le L^* + \epsilon. 
    \end{align*}
\end{lemma}

\subsection{Global Convergence of the Zero-th Order Algorithm}\label{sec:global_convergence}

In this section, we provide the following global convergence theorem. 

\begin{theorem}[Global convergence, formal version of Theorem~\ref{thm:global_conv_informal}]
\label{thm:global_conv_formal}
Given
\begin{itemize}
    \item Let $A_j \in \R^{n \times d}$, $b_j \in \R^n$ satisfies $\| b_j \|_1 \leq 1$ for $\forall j \in [n]$
    \item Let $R \geq 4$, $\| A_j \| \leq R$, $\| x \|_2 \leq R$
    \item Let $W = \diag(w)$, where $\min_i w_i^2 \geq {\mu}/{\sigma_{\min}(A_j)}$ for all $j \in [n]$
    \item batch size $|\mathcal{B}| = B$
    \item Let let $\kappa(A)=\max_{j\in [n]} \kappa(A_j)$
    \item Let $x_0$ denote the initial point
    \item Let $L(x)$ be defined as Definition~\ref{def:softmax_loss}
    \item Let $L^* = \min_{x} L(x)$
    \item Let $M := \exp(O(R^2 + \log n))$
    \item Let 
    \begin{align*}
        t = O( M \cdot(1 + d^{1.5} \cdot \kappa^2(A) / k  ) \cdot   \mu^{-2} B^{-1} \log ( ( L(x_0) - L^*) / \epsilon ) ). 
    \end{align*}
    \end{itemize}
    We perform GD algorithm based on zero-th order (Definition~\ref{def:gd_step}) gradient estimate on it. Then after $t$ 
    iterations, we have
    \begin{align*}
        \E[L(x_t)] \le L^* + \epsilon. 
    \end{align*}
\end{theorem}

\begin{proof}
    Using Corollary~\ref{cor:zo_and_sgd}, we obtain
    \begin{align*}
        \E[L(x_{t+1} ) |x_t] - L(x_t) \le \frac{1}{\gamma} \cdot [-\eta_0 \|\nabla L(x_t)\|^2 + \frac{1}{2} \eta_0^2 \ell \cdot \E[\|\nabla L(x; \mathcal{B})\|^2]],
    \end{align*}
    where $\eta_0$ is the learning rate used in traditional SGD. Note that 
    \begin{align*}
        \E[\|\nabla L(x_t; \mathcal{B})\|^2] = \|\nabla L(x_t)\|^2 + \frac{1}{B}\tr[\Sigma(x_t)].
    \end{align*}
    This follows from the definition of $\Sigma(x)$ (Definition~\ref{def:grad_covv}). 
    
    By selecting $\eta_0 \le \frac{1}{l}$, we have
    \begin{align*}
        \E[L(x_{t+1} ) |x_t] - L(x_t) \le \frac{1}{\gamma} \cdot (-\frac{\eta_0}{2}\|\nabla L(x_t)\|^2+\frac{\eta_0^2 l}{2B}\tr[\Sigma(x_t)]). 
    \end{align*}
    
    By Lemma~\ref{lem:softmax_convex} and Lemma~\ref{lem:softmax_upper_bound_covv}, we have
    \begin{align*}
        \E[L(x_{t+1} ) |x_t] - L(x_t) \le \frac{1}{\gamma}(-\eta_0 \mu + \frac{\eta_0^2 l \alpha}{2B}) \cdot (\E[L(x_t)] - L^*). 
    \end{align*}

    
    Thus by simple algebra, we obtain
    \begin{align*}
        \E[L(x_{t+1})] - L^*\le (1 - \frac{1}{\gamma}(\eta_0\mu - \frac{\eta_0^2 l \alpha}{2B}))\cdot (\E[L(x_t)] - L^*). 
    \end{align*}
    
    Now by choosing $\eta_0 = \min\{\frac{1}{l}, \frac{\mu B}{l\alpha}\}$, we have
    \begin{align*}
        \E[L(x_{t+1})] - L^* \le (1 - \frac{1}{\gamma} \cdot \min\{\frac{\mu}{2l}, \frac{\mu^2B}{2l\alpha}\})(\E[L(x_t)] - L^*).
    \end{align*}
    
    Now, to make $\E[L(x_{t})] - L^* \le \epsilon$, we need 
    \begin{align*}
        t = \gamma \max\{\frac{2l}{\mu}, \frac{2l\alpha}{\mu^2 B}\}\log\frac{L(x_0) - L^*}{\epsilon} 
    \end{align*}
    iterations. 

    Plugging $\gamma$ and $l$, we get
    \begin{align*}
        t = & ~ 16RR_f \cdot(\frac{d^2 \cdot \sqrt{2d+2} \cdot \kappa^2(A) + d - 2}{k(d + 2)} + 1) \cdot \max\{\frac{1}{\mu}, \frac{\alpha}{\mu^2 B}\}\log\frac{L(x_0) - L^*}{\epsilon} \\
        = & ~ 16 R \beta^{-2} n^{1.5} \exp(3 R^2) \cdot(\frac{d^2 \cdot \sqrt{2d+2} \cdot \kappa^2(A) + d - 2}{k(d + 2)} + 1) \cdot \max\{\frac{1}{\mu}, \frac{\alpha}{\mu^2 B}\}\log\frac{L(x_0) - L^*}{\epsilon} \\
        = & ~ 16 R  n^{1.5} \exp(5 R^2) \cdot(\frac{d^2 \cdot \sqrt{2d+2} \cdot \kappa^2(A) + d - 2}{k(d + 2)} + 1) \cdot \max\{\frac{1}{\mu}, \frac{\alpha}{\mu^2 B}\}\log\frac{L(x_0) - L^*}{\epsilon} \\
        = & ~ O(n^{1.5} \exp(30 R^2)  \cdot(\frac{d^2 \cdot \sqrt{2d+2} \cdot \kappa^2(A) + d - 2}{k(d + 2)} + 1) \cdot \mu^{-2}B^{-1}\log\frac{L(x_0) - L^*}{\epsilon}) \\
        = &~ O( M \cdot(1 + d^{1.5} \cdot \kappa^2(A) / k  ) \cdot   \mu^{-2} B^{-1} \log ( ( L(x_0) - L^*) / \epsilon ) )
        ,
    \end{align*}
    where step 1 follows from plugging $\gamma$ and $l$, step 2 follows from plugging $R_f$(Fact~\ref{fac:dls23_parameters}), step 3 follows from plugging $\beta$ (Lemma~\ref{lem:beta}), step 4 follows from $R \ge 4$ and the choosing $\alpha$ to be a large constant, step 5 follows from the definition of $M$. 

    Thus we complete the proof. 
\end{proof}

\ifdefined\isarxiv
\bibliographystyle{alpha}
\bibliography{ref}
\else
\bibliography{ref}

\newcommand{\etalchar}[1]{$^{#1}$}
\begin{thebibliography}{WYW{\etalchar{+}}23}

\bibitem[AS23]{as23}
Josh Alman and Zhao Song.
\newblock Fast attention requires bounded entries.
\newblock {\em arXiv preprint arXiv:2302.13214}, 2023.

\bibitem[BCE{\etalchar{+}}23]{bce+23}
S{\'e}bastien Bubeck, Varun Chandrasekaran, Ronen Eldan, Johannes Gehrke, Eric
  Horvitz, Ece Kamar, Peter Lee, Yin~Tat Lee, Yuanzhi Li, Scott Lundberg,
  et~al.
\newblock Sparks of artificial general intelligence: Early experiments with
  gpt-4.
\newblock {\em arXiv preprint arXiv:2303.12712}, 2023.

\bibitem[BG22]{bg22}
Krishnakumar Balasubramanian and Saeed Ghadimi.
\newblock Zeroth-order nonconvex stochastic optimization: Handling constraints,
  high dimensionality, and saddle points.
\newblock {\em Foundations of Computational Mathematics}, pages 1--42, 2022.

\bibitem[BMR{\etalchar{+}}20]{bmr+20}
Tom Brown, Benjamin Mann, Nick Ryder, Melanie Subbiah, Jared~D Kaplan, Prafulla
  Dhariwal, Arvind Neelakantan, Pranav Shyam, Girish Sastry, Amanda Askell,
  et~al.
\newblock Language models are few-shot learners.
\newblock {\em Advances in neural information processing systems},
  33:1877--1901, 2020.

\bibitem[BSZ23]{bsz23}
Jan van~den Brand, Zhao Song, and Tianyi Zhou.
\newblock Algorithm and hardness for dynamic attention maintenance in large
  language models.
\newblock {\em arXiv preprint arXiv:2304.02207}, 2023.

\bibitem[Cha22]{cha22}
ChatGPT.
\newblock Optimizing language models for dialogue.
\newblock {\em OpenAI Blog}, November 2022.

\bibitem[CLP{\etalchar{+}}21]{clp+21}
Beidi Chen, Zichang Liu, Binghui Peng, Zhaozhuo Xu, Jonathan~Lingjie Li, Tri
  Dao, Zhao Song, Anshumali Shrivastava, and Christopher Re.
\newblock Mongoose: A learnable lsh framework for efficient neural network
  training.
\newblock In {\em International Conference on Learning Representations}, 2021.

\bibitem[CND{\etalchar{+}}22]{cnd+22}
Aakanksha Chowdhery, Sharan Narang, Jacob Devlin, Maarten Bosma, Gaurav Mishra,
  Adam Roberts, Paul Barham, Hyung~Won Chung, Charles Sutton, Sebastian
  Gehrmann, et~al.
\newblock Palm: Scaling language modeling with pathways.
\newblock {\em arXiv preprint arXiv:2204.02311}, 2022.

\bibitem[CNW15]{cnw15}
Michael~B Cohen, Jelani Nelson, and David~P Woodruff.
\newblock Optimal approximate matrix product in terms of stable rank.
\newblock {\em arXiv preprint arXiv:1507.02268}, 2015.

\bibitem[CZS{\etalchar{+}}17]{czs+17}
Pin-Yu Chen, Huan Zhang, Yash Sharma, Jinfeng Yi, and Cho-Jui Hsieh.
\newblock Zoo: Zeroth order optimization based black-box attacks to deep neural
  networks without training substitute models.
\newblock In {\em Proceedings of the 10th ACM workshop on artificial
  intelligence and security}, pages 15--26, 2017.

\bibitem[DCLT18]{dclt18}
Jacob Devlin, Ming-Wei Chang, Kenton Lee, and Kristina Toutanova.
\newblock Bert: Pre-training of deep bidirectional transformers for language
  understanding.
\newblock {\em arXiv preprint arXiv:1810.04805}, 2018.

\bibitem[DJWW15]{djww15}
John~C Duchi, Michael~I Jordan, Martin~J Wainwright, and Andre Wibisono.
\newblock Optimal rates for zero-order convex optimization: The power of two
  function evaluations.
\newblock {\em IEEE Transactions on Information Theory}, 61(5):2788--2806,
  2015.

\bibitem[DLS23]{dls23}
Yichuan Deng, Zhihang Li, and Zhao Song.
\newblock Attention scheme inspired softmax regression.
\newblock {\em arXiv preprint arXiv:2304.10411}, 2023.

\bibitem[DMS23]{dms23}
Yichuan Deng, Sridhar Mahadevan, and Zhao Song.
\newblock Randomized and deterministic attention sparsification algorithms for
  over-parameterized feature dimension.
\newblock {\em arxiv preprint: arxiv 2304.03426}, 2023.

\bibitem[GKK{\etalchar{+}}19]{gkk+19}
Daniel Golovin, John Karro, Greg Kochanski, Chansoo Lee, Xingyou Song, and
  Qiuyi Zhang.
\newblock Gradientless descent: High-dimensional zeroth-order optimization.
\newblock {\em arXiv preprint arXiv:1911.06317}, 2019.

\bibitem[GL13]{gl13}
Saeed Ghadimi and Guanghui Lan.
\newblock Stochastic first-and zeroth-order methods for nonconvex stochastic
  programming.
\newblock {\em SIAM Journal on Optimization}, 23(4):2341--2368, 2013.

\bibitem[GMS23]{gms23}
Yeqi Gao, Sridhar Mahadevan, and Zhao Song.
\newblock An over-parameterized exponential regression.
\newblock {\em arXiv preprint arXiv:2303.16504}, 2023.

\bibitem[GSX23]{gsx23}
Yeqi Gao, Zhao Song, and Shenghao Xie.
\newblock In-context learning for attention scheme: from single softmax
  regression to multiple softmax regression via a tensor trick.
\newblock {\em arXiv preprint arXiv:2307.02419}, 2023.

\bibitem[GSY23a]{gsy23_dp}
Yeqi Gao, Zhao Song, and Xin Yang.
\newblock Differentially private attention computation.
\newblock {\em arXiv preprint arXiv:2305.04701}, 2023.

\bibitem[GSY23b]{gsy23_hyper}
Yeqi Gao, Zhao Song, and Junze Yin.
\newblock An iterative algorithm for rescaled hyperbolic functions regression.
\newblock {\em arXiv preprint arXiv:2305.00660}, 2023.

\bibitem[Hin22]{h22}
Geoffrey Hinton.
\newblock The forward-forward algorithm: Some preliminary investigations.
\newblock {\em arXiv preprint arXiv:2212.13345}, 2022.

\bibitem[HWL21]{hwl21}
Weihua He, Yongyun Wu, and Xiaohua Li.
\newblock Attention mechanism for neural machine translation: A survey.
\newblock In {\em 2021 IEEE 5th Information Technology, Networking, Electronic
  and Automation Control Conference (ITNEC)}, volume~5, pages 1485--1489. IEEE,
  2021.

\bibitem[KKL20]{kkl20}
Nikita Kitaev, {\L}ukasz Kaiser, and Anselm Levskaya.
\newblock Reformer: The efficient transformer.
\newblock {\em arXiv preprint arXiv:2001.04451}, 2020.

\bibitem[KNS16]{kns16}
Hamed Karimi, Julie Nutini, and Mark Schmidt.
\newblock Linear convergence of gradient and proximal-gradient methods under
  the polyak-{\l}ojasiewicz condition.
\newblock In {\em Machine Learning and Knowledge Discovery in Databases:
  European Conference, ECML PKDD 2016, Riva del Garda, Italy, September 19-23,
  2016, Proceedings, Part I 16}, pages 795--811. Springer, 2016.

\bibitem[KW52]{kw52}
Jack Kiefer and Jacob Wolfowitz.
\newblock Stochastic estimation of the maximum of a regression function.
\newblock {\em The Annals of Mathematical Statistics}, pages 462--466, 1952.

\bibitem[LCCH18]{lcch18}
Sijia Liu, Pin-Yu Chen, Xiangyi Chen, and Mingyi Hong.
\newblock signsgd via zeroth-order oracle.
\newblock In {\em International Conference on Learning Representations}, 2018.

\bibitem[LLR23]{llr23}
Yuchen Li, Yuanzhi Li, and Andrej Risteski.
\newblock How do transformers learn topic structure: Towards a mechanistic
  understanding.
\newblock {\em arXiv preprint arXiv:2303.04245}, 2023.

\bibitem[LRV{\etalchar{+}}20]{lrv+20}
Sijia Liu, Parikshit Ram, Deepak Vijaykeerthy, Djallel Bouneffouf, Gregory
  Bramble, Horst Samulowitz, Dakuo Wang, Andrew Conn, and Alexander Gray.
\newblock An admm based framework for automl pipeline configuration.
\newblock In {\em Proceedings of the AAAI Conference on Artificial
  Intelligence}, volume~34, pages 4892--4899, 2020.

\bibitem[LSX{\etalchar{+}}23]{lsx+23}
Shuai Li, Zhao Song, Yu~Xia, Tong Yu, and Tianyi Zhou.
\newblock The closeness of in-context learning and weight shifting for softmax
  regression.
\newblock {\em arXiv preprint arXiv:2304.13276}, 2023.

\bibitem[LSZ23]{lsz23}
Zhihang Li, Zhao Song, and Tianyi Zhou.
\newblock Solving regularized exp, cosh and sinh regression problems.
\newblock {\em arXiv preprint, 2303.15725}, 2023.

\bibitem[MGN{\etalchar{+}}23]{mgn+23}
Sadhika Malladi, Tianyu Gao, Eshaan Nichani, Alex Damian, Jason~D Lee, Danqi
  Chen, and Sanjeev Arora.
\newblock Fine-tuning language models with just forward passes.
\newblock {\em arXiv preprint arXiv:2305.17333}, 2023.

\bibitem[MGR18]{mgr18}
Horia Mania, Aurelia Guy, and Benjamin Recht.
\newblock Simple random search of static linear policies is competitive for
  reinforcement learning.
\newblock {\em Advances in Neural Information Processing Systems}, 31, 2018.

\bibitem[MMS{\etalchar{+}}19]{mms+19}
Louis Martin, Benjamin Muller, Pedro Javier~Ortiz Suarez, Yoann Dupont, Laurent
  Romary, Eric~Villemonte de~La~Clergerie, Djame Seddah, and Benoit Sagot.
\newblock Camembert: a tasty french language model.
\newblock {\em arXiv preprint arXiv:1911.03894}, 2019.

\bibitem[NM65]{nm65}
John~A Nelder and Roger Mead.
\newblock A simplex method for function minimization.
\newblock {\em The computer journal}, 7(4):308--313, 1965.

\bibitem[NS17]{ns17}
Yurii Nesterov and Vladimir Spokoiny.
\newblock Random gradient-free minimization of convex functions.
\newblock {\em Foundations of Computational Mathematics}, 17:527--566, 2017.

\bibitem[Ope23]{o23}
OpenAI.
\newblock Gpt-4 technical report.
\newblock {\em arXiv preprint arXiv:2303.08774}, 2023.

\bibitem[RNS{\etalchar{+}}18]{rns+18}
Alec Radford, Karthik Narasimhan, Tim Salimans, Ilya Sutskever, et~al.
\newblock Improving language understanding by generative pre-training.
\newblock {\em .}, 2018.

\bibitem[RWC{\etalchar{+}}19]{rwc+19}
Alec Radford, Jeffrey Wu, Rewon Child, David Luan, Dario Amodei, Ilya
  Sutskever, et~al.
\newblock Language models are unsupervised multitask learners.
\newblock {\em OpenAI blog}, 1(8):9, 2019.

\bibitem[SHC{\etalchar{+}}17]{shc17}
Tim Salimans, Jonathan Ho, Xi~Chen, Szymon Sidor, and Ilya Sutskever.
\newblock Evolution strategies as a scalable alternative to reinforcement
  learning.
\newblock {\em arXiv preprint arXiv:1703.03864}, 2017.

\bibitem[SLA12]{sla12}
Jasper Snoek, Hugo Larochelle, and Ryan~P Adams.
\newblock Practical bayesian optimization of machine learning algorithms.
\newblock {\em Advances in neural information processing systems}, 25, 2012.

\bibitem[Spa87]{spa87}
James~C Spall.
\newblock A stochastic approximation technique for generating maximum
  likelihood parameter estimates.
\newblock In {\em 1987 American control conference}, pages 1161--1167. IEEE,
  1987.

\bibitem[Spa92]{Spa92}
James~C Spall.
\newblock Multivariate stochastic approximation using a simultaneous
  perturbation gradient approximation.
\newblock {\em IEEE transactions on automatic control}, 37(3):332--341, 1992.

\bibitem[Spa98]{spa98}
James~C Spall.
\newblock Implementation of the simultaneous perturbation algorithm for
  stochastic optimization.
\newblock {\em IEEE Transactions on aerospace and electronic systems},
  34(3):817--823, 1998.

\bibitem[SSZ23]{ssz23}
Ritwik Sinha, Zhao Song, and Tianyi Zhou.
\newblock A mathematical abstraction for balancing the trade-off between
  creativity and reality in large language models.
\newblock {\em arXiv preprint arXiv:2306.02295}, 2023.

\bibitem[SZKS21]{szks21}
Charlie Snell, Ruiqi Zhong, Dan Klein, and Jacob Steinhardt.
\newblock Approximating how single head attention learns.
\newblock {\em arXiv preprint arXiv:2103.07601}, 2021.

\bibitem[UAS{\etalchar{+}}20]{uas20}
Mohd Usama, Belal Ahmad, Enmin Song, M~Shamim Hossain, Mubarak Alrashoud, and
  Ghulam Muhammad.
\newblock Attention-based sentiment analysis using convolutional and recurrent
  neural network.
\newblock {\em Future Generation Computer Systems}, 113:571--578, 2020.

\bibitem[VSP{\etalchar{+}}17]{vsp+17}
Ashish Vaswani, Noam Shazeer, Niki Parmar, Jakob Uszkoreit, Llion Jones,
  Aidan~N Gomez, {\L}ukasz Kaiser, and Illia Polosukhin.
\newblock Attention is all you need.
\newblock {\em Advances in neural information processing systems}, 30, 2017.

\bibitem[WYW{\etalchar{+}}23]{wyw+23}
Junda Wu, Tong Yu, Rui Wang, Zhao Song, Ruiyi Zhang, Handong Zhao, Chaochao Lu,
  Shuai Li, and Ricardo Henao.
\newblock Infoprompt: Information-theoretic soft prompt tuning for natural
  language understanding.
\newblock {\em arXiv preprint arXiv:2306.04933}, 2023.

\bibitem[ZHDK23]{zhdk23}
Amir Zandieh, Insu Han, Majid Daliri, and Amin Karbasi.
\newblock Kdeformer: Accelerating transformers via kernel density estimation.
\newblock {\em arXiv preprint arXiv:2302.02451}, 2023.

\bibitem[ZHL{\etalchar{+}}23]{zhl+23}
Eric Zelikman, Qian Huang, Percy Liang, Nick Haber, and Noah~D Goodman.
\newblock Just one byte (per gradient): A note on low-bandwidth decentralized
  language model finetuning using shared randomness.
\newblock {\em arXiv preprint arXiv:2306.10015}, 2023.

\bibitem[ZRG{\etalchar{+}}22]{zrg+22}
Susan Zhang, Stephen Roller, Naman Goyal, Mikel Artetxe, Moya Chen, Shuohui
  Chen, Christopher Dewan, Mona Diab, Xian Li, Xi~Victoria Lin, et~al.
\newblock Opt: Open pre-trained transformer language models.
\newblock {\em arXiv preprint arXiv:2205.01068}, 2022.

\bibitem[ZSZ{\etalchar{+}}23]{zsz+23}
Zhenyu Zhang, Ying Sheng, Tianyi Zhou, Tianlong Chen, Lianmin Zheng, Ruisi Cai,
  Zhao Song, Yuandong Tian, Christopher R{\'{e}}, Clark~W. Barrett, Zhangyang
  Wang, and Beidi Chen.
\newblock H2o: Heavy-hitter oracle for efficient generative inference of large
  language models.
\newblock {\em CoRR}, abs/2306.14048, 2023.

\end{thebibliography}
\bibliographystyle{alpha}

\fi

\newpage
\onecolumn
\appendix




\end{document}